\documentclass[10pt,twocolumn,letterpaper]{article}

\usepackage[pagenumbers]{wacv} 

\usepackage{graphicx}
\usepackage{amsmath}
\usepackage[table]{xcolor}
\usepackage{amssymb}
\usepackage{booktabs}
\usepackage{multirow}
\usepackage{array}
\usepackage{enumitem}
\usepackage{tablefootnote, threeparttable, balance, hhline}

\usepackage[pagebackref,breaklinks,colorlinks]{hyperref}

\usepackage[capitalize]{cleveref}
\crefname{section}{Sec.}{Secs.}
\Crefname{section}{Section}{Sections}
\Crefname{table}{Table}{Tables}
\crefname{table}{Tab.}{Tabs.}

\def\wacvPaperID{1603} 
\def\confName{WACV}
\def\confYear{2025}

\definecolor{lightblue}{RGB}{173, 216, 230}
\newcommand{\thickhline}{\noalign{\hrule height 0.4mm}}
\usepackage{fancyhdr,tikz}

\begin{document}

\title{A Parametric Approach to Adversarial Augmentation for\\ Cross-Domain Iris Presentation Attack Detection}

\author{Debasmita Pal, Redwan Sony, Arun Ross\\
Michigan State University,
East Lansing, MI 48824, USA\\
{\tt\small paldebas@msu.edu, sonymd@msu.edu, rossarun@cse.msu.edu}
}
\maketitle
\renewcommand{\thefootnote}{\fnsymbol{footnote}} 
\footnotetext[2]{This version fixes an incorrect entry in Table 5 published in the WACV 2025 paper. The correction does not affect the findings or conclusions of this work.}
\renewcommand{\thefootnote}{\arabic{footnote}}   
\begin{tikzpicture}[remember picture, overlay]
\node[anchor=north, text width=\paperwidth, align=center] at (current page.north) [shift={(0,-1.5cm)}] {\textcolor{red}{Accepted to the IEEE/CVF Winter Conference on Applications of Computer Vision (WACV), Arizona, USA, 2025.}
};
\end{tikzpicture}

\vspace{-\baselineskip}
\vspace{-\baselineskip}
\begin{abstract}
Iris-based biometric systems are vulnerable to presentation attacks (PAs), where adversaries present physical artifacts (e.g., printed iris images, textured contact lenses) to defeat the system. This has led to the development of various presentation attack detection (PAD) algorithms, which typically perform well in intra-domain settings. However, they often struggle to generalize effectively in cross-domain scenarios, where training and testing employ different sensors, PA instruments, and datasets. In this work, we use adversarial training samples of both bonafide irides and PAs to improve the cross-domain performance of a PAD classifier. The novelty of our approach lies in leveraging transformation parameters from classical data augmentation schemes (e.g., translation, rotation) to generate adversarial samples. We achieve this through a convolutional autoencoder, ADV-GEN, that inputs original training samples along with a set of geometric and photometric transformations. The transformation parameters act as regularization variables, guiding ADV-GEN to generate adversarial samples in a constrained search space. Experiments conducted on the LivDet-Iris 2017 database, comprising four datasets, and the LivDet-Iris 2020 dataset, demonstrate the efficacy of our proposed method. The code is available at {\small\url{https://github.com/iPRoBe-lab/ADV-GEN-IrisPAD}}. 
\end{abstract}
\vspace{-\baselineskip}
\section{Introduction}
\label{sec:Introduction}
Iris recognition systems, which use the intricate textural pattern of the iris as a biometric cue \cite{Daugman1993}, have demonstrated very high recognition accuracy in constrained settings.\footnote{\url{https://www.nist.gov/programs-projects/iris-exchange-irex-overview}} However, these systems are vulnerable to adversarial attacks, including presentation attacks (PAs), where an adversary presents a physical artifact (\eg, printed iris image, cosmetic contact lens, prosthetic eye) or a modified iris (\eg, surgery) to spoof an identity, conceal one's own identity or create a virtual identity to evade the system. 

To counteract this, numerous presentation attack detection (PAD) schemes have been developed. These schemes have formulated PAD as a binary classification task to distinguish between bonafides (real irides) and PAs. Specifically, deep neural network (DNN)-based approaches have gained popularity due to their superior performance over traditional machine learning techniques \cite{Yambay2017}. These methods have proven successful in intra-domain settings, where training and testing datasets exhibit similar characteristics and homogeneous data distributions. However, their performance considerably degrades in cross-domain scenarios where training and testing are conducted on different datasets involving previously unseen PAs, different sensors, and varying acquisition environments. A recent comprehensive study on open-set iris PAD by Boyd \etal \cite{Boyd2023}, underscored the necessity for continued research to enhance the generalizability of PAD algorithms. 

This work aims to improve the generalizability of a DNN-based iris PAD classifier by invoking the principle of adversarial augmentation. Adversarial images are input samples that have been subtly perturbed by the injection of carefully crafted noise, forcing the classifier to produce an incorrect output \cite{Szegedy2014}. However, these images, i.e., samples, can also be utilized to enhance a classifier's performance, generalizability and robustness when incorporated into the training set \cite{Goodfellow2015,deepfool2016,Xie2020,Chen2022}. Many studies have shown that adversarial images lead DNNs to learn meaningful and robust feature representations \cite{Tsipras2019,Ilyas2019}. 

In general, the effectiveness of classical data augmentation (DA) techniques such as rotation, scaling, coloring, has been illustrated in the literature to reduce over-fitting and improve model generalization by increasing the diversity of training samples \cite{He2016,Krizhevsky2012}. Recent works have further explored automated search-based augmentation schemes \cite{Cubak2019,Muller2021}. However, researchers argue that while DA techniques are beneficial, they alone may not be sufficient to achieve model robustness; the ultimate goal is to learn domain-invariant features \cite{Yin2019}. 

In this work, we design an adversarial image generator based on a Convolutional Autoencoder (CAE) \cite{autoencoder}, named {\bf Adversarial Generator (ADV-GEN)}, under white-box settings, where the adversary (generator) has access to the classifier -- the iris PAD classifier, in our case. This model inputs the original training images, consisting of both bonafide irides and PAs, along with the geometric and photometric transformation parameters used for augmentation, to synthesize adversarial images for both classes. These adversarial images along with the original training images are then used to retrain the classifier. 


In \cite{Huang2018}, the authors used using a convex constrained optimization approach for generating adversarial iris images to improve the performance of an iris recognition algorithm based on IrisCodes. Classical DA techniques along with fusion-based strategies have been explored to improve iris PAD performance \cite{Fang2022, Tapia2022}. In \cite{Ogin2024}, the authors first constructed a physical impersonation attack using adversarial perturbation on the iris region and a physical PAD evading attack using an adversarial patch on the pupil region. Then they used adversarial fine-tuning to counter both attacks. 

Our approach, on the other hand, focuses on generating adversarial images by leveraging the {\it parameter space} of typical  geometric and photometric transformations used for data augmentation. This is in contrast to traditional adversarial image generation methods that inject noise to induce perturbation along the gradients of the loss function with respect to the input image \cite{Szegedy2014, Goodfellow2015}.


\section{Related Work}
\label{sec:RelatedWork}
\subsection{Iris Presentation Attack Detection (PAD)}
PAD methods can be hardware-based (sensor-based) or software-based (image-based). Hardware-based PAD methods leverage additional sensors or devices alongside iris sensors to facilitate PA detection (also known as liveness detection), while software-based solutions analyze digital iris images captured by iris sensors to extract salient features for determining if an image is a bonafide or a PA. 

Examples of hardware-based approaches include \cite{Czajka2016, Hughes2013, Connell2013, Czajka2019, Komogortsev2013, Czajka2023}. 
In software-based approaches, which is the focus of our work, techniques such as local binary pattern (LBP), local phase quantization (LPQ), binarized statistical image features (BSIF), and shift-invariant descriptors (SID) have been employed \cite{Gragnaniello2015, Hu2016, Doyle2015, Raghavendra2015}. With the success of DNNs in various computer vision tasks, researchers have also explored Convolution Neural Network (CNN)-based classifiers such as {\em spoofnet} \cite{Menotti2015}, multi-patch CNN \cite{He2016PAD} and triplet CNN-based framework \cite{Pala2017}. Yadav \etal proposed fusing multi-level Haralick texture features (handcrafted) with VGG features (DNN-based) to capture textural variations in iris images \cite{Yadav2018}. Kuehlkamp \etal proposed an ensemble learning technique using multiple CNNs trained on pre-trained BSIF filters \cite{Kuehlkamp2019}. Fang \etal employed a fusion-based classifier combining 2D textural iris features and 3D photometric stereo shape features \cite{Fang2021}. Chen and Ross presented a multi-task CNN approach capable of simultaneously performing iris localization and PA detection \cite{Chen2018}. 

Yadav \etal proposed DensePAD, using a 22-layer DenseNet architecture to detect textured contact lenses \cite{Yadav2019}. In \cite{Sharma2020, Das2020}, D-NetPAD using the DenseNet-121 and DenseNet-161 architectures was developed, achieving good results in PA detection. Sharma \etal later examined the sensitivity of weight perturbations in DNNs, including D-NetPAD, and proposed an ensemble of perturbed models to improve PAD performance over the the original models \cite{Sharma2024}. Chen and Ross introduced attention guided iris PAD incorporating attention modules on top of the last convolutional layer of the DenseNet-121 \cite{Chen2021}. Swarup \etal introduced a convolution block attention mechanism between the blocks of the DenseNet-201 \cite{Swarup2022}.   

Hoffman \etal adopted a patch-batch training approach for CNNs to enhance detection accuracy and generalizability across different datasets and sensors \cite{Hoffman2018}. In \cite{Hoffman2019}, they extended by integrating outputs from three CNNs, each analyzing different regions: the iris, the entire ocular region, and a subset of sampled patches from the ocular region. Fang \etal proposed a micro stripes analysis (MSA) framework to learn image dynamics around the iris/sclera boundary area, analyzing its significance in cross-database and unknown attack evaluation \cite{Fang2021MSA}. Agarwal \etal developed a deep CNN-based approach using early and late fusion of various iris images and scores for contact lens detection \cite{Agarwal2022TBIOM}. Gupta \etal introduced a generalized DNN-based PAD framework, MVANet \cite{Gupta2021}. Another study by Agarwal \etal presented a two head contraction-expansion CNN utilizing raw iris images and edge-enhanced iris images \cite{Agarwal2022PRL}. Yadav \etal used the Relativistic Average Standard Generative Adversarial Network (RaSGAN) to generate synthetic irides, improving PAD performance for unseen attacks \cite{Yadav2019RasGAN}. Jain \etal introduced a multi-task convolutional skip autoencoder-based framework performing denoising and classification together \cite{Jain2022}. Fang \etal proposed an attention-based deep pixel-wise binary supervision (A-PBS) method, improving both intra- and cross-spectrum iris PAD performance \cite{Fang2023}. Li \etal suggested a Single Domain Dynamic Generalization (SDDG) framework incorporating domain-specific as well as domain-invariant features \cite{Li2023}. In addition, a series of Liveness Detection (LivDet)-Iris competitions have been launched to advance research on iris PAD \cite{Yambay2017, Das2020, Yambay2023, Tinsley2023}.    

\subsection{Adversarial Image Generation and DA}

Several algorithms have been proposed to generate adversarial images, including the Fast Sign Gradient Method (FGSM) \cite{Goodfellow2015}, basic iterative method and iterative least-likely class method \cite{Kurakin2017}, optimization-based methods \cite{Carlini2017}, DeepFool algorithm \cite{deepfool2016}. Further, these adversarial samples have been used for training to enhance model robustness against adversarial attacks \cite{Goodfellow2015, Kurakin2017, deepfool2016}. Baluja and Fischer introduced the Adversarial Transformation Network to generate adversarial images using a feed-forward neural network \cite{Baluja2018}. Xiao \etal proposed ADV-GAN based on Generative Adversarial Network (GAN) \cite{Xiao2018}.

A comprehensive survey on various DA techniques, including AutoDA, is presented in \cite{Shorten2019, Yang2023}. Zhang \etal developed an adversarial method for augmentation policy search during network training to enhance robust feature learning \cite{Zhang2020}. Further, generative models such as GANs \cite{GAN2020} have been widely used for DA. Recently, Trabucco \etal utilized text-to-image diffusion models (DA-Fusion) to generate new visual concepts, addressing the lack of diversity in DA techniques \cite{Trabucco2024}.     

\section{Proposed Method}
\label{sec:Methodology}
Aiming to improve the performance of a PAD classifier across different domains/datasets, we implement our proposed method using training images from a single domain ($D_0$), comprising $m$ bonafides and $n$ PAs. Initially, we detect the irides from the original images from both bonafides and PAs, using the VeriEye iris-segmenter.\footnote{\url{www.neurotechnology.com/verieye.html}} The segmented irides are then center-cropped and resized to 224$\times$224 pixels, serving as the training images, $X= \{x_1,x_2,..,x_{m+n}\}$, with corresponding labels, $Y=\{y_1, y_2, ..., y_{m+n}\}$, where $y_i \in \{0, 1\}$ (0: bonafide and 1: PA).  The proposed approach then proceeds as follows:
\begin{itemize}[topsep=0em, noitemsep,leftmargin=*]
    \item Train the \textbf{Standard PAD classifier} ($F$) using $(X,Y)$; $F(x_i)=\hat{y_i}\in[0,1]$. 
    \item Train the CAE-based \textbf{Adversarial Generator (ADV-GEN)} ($G$) using $X$ along with a set of randomly sampled transformation parameters, $T=\{t_1, t_2, ...t_{m+n}$) (see Table \ref{table:transformation}), such that the generated output images, $X'=\{x_1', x_2', ...x_{m+n}'\}$, visually resemble $t_i \cdot x_i$ (images produced after applying $t_i$ to $x_i$), but are incorrectly classified by $F$. So, $G(x_i,t_i)=x_i'$ such that $( \|t_i \cdot x_i - x_i'\|_2 )\to 0.0$ and  $F(x_i')\to(1-y_i)$.
    \item Generate adversarial images for each $x_i$ using the trained CAE, resulting in $m$ adversarial bonafides (bonafides classified as PAs) and $n$ adversarial PAs (PAs classified as bonafides).
    \item Select a subset of adversarial PA samples ($P$) and adversarial bonafide ($Q$) samples from $X'$ using squared L2-norm and K-means clustering to ensure both similarity with the transformed original images ($t_i \cdot x_i$) and diversity among the selected images.
    \item Train the \textbf{Adversarially Augmented PAD (AA-PAD)} classifier ($F'$) with ($X\cup P\cup Q$). 
    \item Evaluate the AA-PAD classifier on the test samples from other domains different from $D_0$.
\end{itemize}

\begin{table}[!h]
    \centering
    \caption{Transformations parameters (TP) and defined ranges used by ADV-GEN (SS: Step Size). An example of a transformation vector is $(-10, 12, -4, -2, 3, 0.8, 1.1, 1.0, 0.9, 0.7)^t$.}
    \vspace{-0.5\baselineskip}
    \begin{tabular}{|>{\centering\arraybackslash}m{1.4cm}|>{\centering\arraybackslash}m{1.9cm}|>{\centering\arraybackslash}m{1.4cm}|>{\centering\arraybackslash}m{1.9cm}|}
         \hline
         {\cellcolor{lightgray}\bf TP} & {\cellcolor{lightgray}\bf Range, SS} & {\cellcolor{lightgray}\bf TP} & {\cellcolor{lightgray}\bf Range, SS} \\
         \thickhline
         translate\_x & [-15, 15], 1.0 &  solarize & [0.7,1.0], 0.1\\
        translate\_y & [-15, 15], 1.0 & scale & [0.9,1.2], 0.1 \\
        shear\_x & [-5, 5], 1.0 & sharpness & [0.5,1.5], 0.1\\
        shear\_y & [-5, 5], 1.0 & brightness & [0.5,1.5], 0.1\\
        rotation & [-10, 10], 1.0 & contrast & [0.5,1.5], 0.1\\
        \hline
    \end{tabular}
    \label{table:transformation}
    \vspace{-\baselineskip}
\end{table}

\subsection{Standard PAD Classifier}

\begin{figure}[t]
\begin{center}
    		\fbox{
			\parbox[][0.21\textheight][c]{0.42\textwidth}{
				\includegraphics[width=0.42\textwidth, height=0.21\textheight]{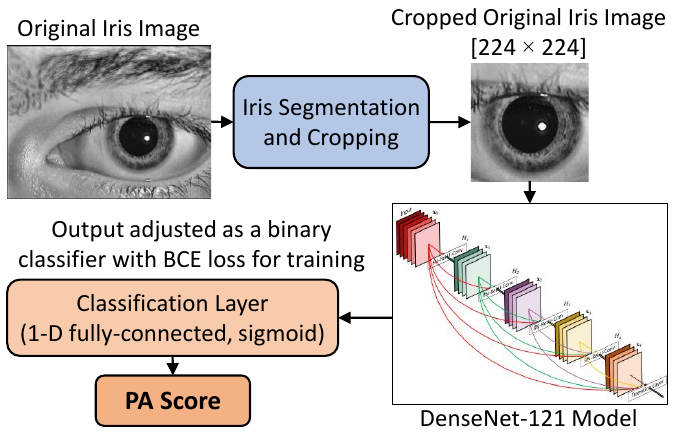}
		}}
  \end{center}
\vspace{-\baselineskip}
  \caption{Training the Standard PAD classifier that uses the DenseNet-121 backbone and inputs a cropped original iris image and outputs a PA score $\in[0,1]$; here, $0$ is the ideal score for a bonafide and $1$ is the ideal score for a PA.}
		\label{fig:PAD_training}
  \vspace{-\baselineskip}
  \end{figure}

We select DenseNet-121 \cite{Huang2017} as the backbone of our PAD classifier due to its architectural advantages and superior performance in PA detection compared to other architectures \cite{Sharma2020}. The pre-trained DenseNet-121 model is fine-tuned as a binary classifier, with one output in the classifier layer, which produces a single score. This score, processed through a sigmoid activation function, represents the probability of the given input image being a PA. A score close to 1 indicates that the input image is a PA, while a score close to 0 indicates a bonafide image. Figure \ref{fig:PAD_training} illustrates the training process of our Standard PAD classifier, where cropped original iris images are used as input. Training is conducted using the binary cross-entropy (BCE) loss function (Eq. \ref{eq:BCE}) and the Adam optimizer with a learning rate of 0.0001 for 50 epochs, selecting the best model based on the validation loss. 
\vspace{-0.5\baselineskip}
\begin{equation}
\label{eq:BCE}
    \mathcal{L}_{BCE}=-\mathbb{E}\left[ y\log(\hat{y}) + (1 - y)\log(1 - \hat{y}) \right]
\vspace{-0.5\baselineskip}
\end{equation}
where, \( \mathcal{L}_{BCE} \): loss function; \( y \): ground truth label of input image; \( \hat{y} \): output PA score by the Standard PAD classifier.

\subsection{ADV-GEN: Generating Adversarial Images}
\begin{figure}
\begin{center}
		\fbox{
			\parbox[][0.16\textheight][c]{0.44\textwidth}{
				\includegraphics[width=0.44\textwidth, height=0.16\textheight]{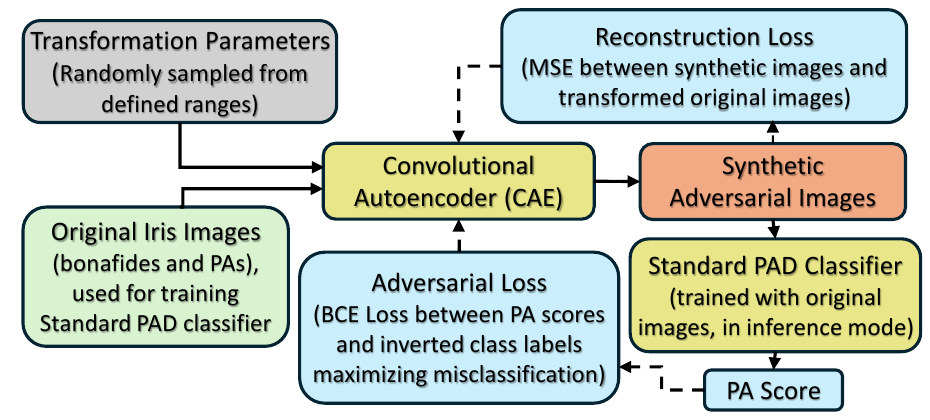}
		}}    
\end{center}
\vspace{-\baselineskip}
	\caption{Schematic of ADV-GEN. The CAE inputs cropped original iris images and a set of transformation parameters to generate synthetic adversarial images using a dual-loss function.}
	\label{fig:architecture}
 \vspace{-\baselineskip}
  \end{figure}

Figure \ref{fig:architecture} outlines the framework of our proposed CAE-based ADV-GEN to generate adversarial images. An autoencoder \cite{autoencoder} is a generative model composed of two components: an encoder and a decoder, typically implemented using neural networks. The encoder $(f)$ maps input data points $s$ (image space) to a feature space $z$ (a compressed representation), while the decoder $(g)$ reconstructs the data points by mapping feature space ($z$) to data space ($s$): $z=f(s)$ and $g(z)=g(f(s))=t$, where $t$ is the reconstruction of $s$. 

The cropped original iris images, both bonafides and PAs, which were used to train the Standard PAD classifier, along with transformation parameter vectors, serve as input to the CAE in the ADV-GEN. We utilize both geometric and photometric transformations (commonly used in DA and sourced from a popular repository), bounded by a defined range of discrete values (Table \ref{table:transformation}) to constrain the parameter space. The ranges of these parameters are selected by trial-and-error on training images from various datasets to ensure the generation of semantically plausible images. During training, randomly sampled values within these ranges are input to the CAE, enabling it to generate transformed iris images, which visually resemble the transformed original images. This generation is guided by a reconstruction loss, viz., Mean Squared Error (MSE), controlling similarity between them. Concurrently, to make these generated transformed iris images adversarial, they are passed through the Standard PAD classifier in inference mode to obtain PA scores. We then backpropagate an adversarial loss (BCE loss between these PA scores and the inverted class labels maximizing the misclassification error) to the CAE, forcing the generated images to be misclassified by the Standard PAD classifier. This dual-loss function approach (Eq. \ref{eq:L_CAE}) enables the effective generation of synthetic adversarial images for both the classes.    
\\
\begin{minipage}{0.49\linewidth}
\vspace{-0.5\baselineskip}
\begin{equation*}
    \mathcal{L}_{CAE} = \mathcal{L}_{res} + \lambda * \mathcal{L}_{adv} 
    \end{equation*}
\end{minipage}
\begin{minipage}{0.49\linewidth}
\begin{equation*}
    \mathcal{L}_{res} = \mathbb{E}[(t\cdot x - x')^2] 
    \end{equation*}
\end{minipage}
\begin{equation}
    \label{eq:L_CAE}
    \mathcal{L}_{adv} = -\mathbb{E} \left[ (1-y) \log(\hat{y}) + y \log(1 - \hat{y}) \right]
\end{equation}
\noindent where, $\lambda$: a tunable hyperparameter to balance between these two losses, set to 0.1 in our experiment, $t$: transformation parameter vector, $x$: original image, $t\cdot x$: transformed original iris image, $x'$: synthetic adversarial image, $y$: ground-truth label of original image, $\hat{y}$: output PA score of synthetic image produced by the Standard PAD classifier.   

\begin{figure}
\begin{center}
    
		\fbox{
			\parbox[][0.35\textheight][c]{0.42\textwidth}{
				\includegraphics[width=0.42\textwidth, height=0.35\textheight]{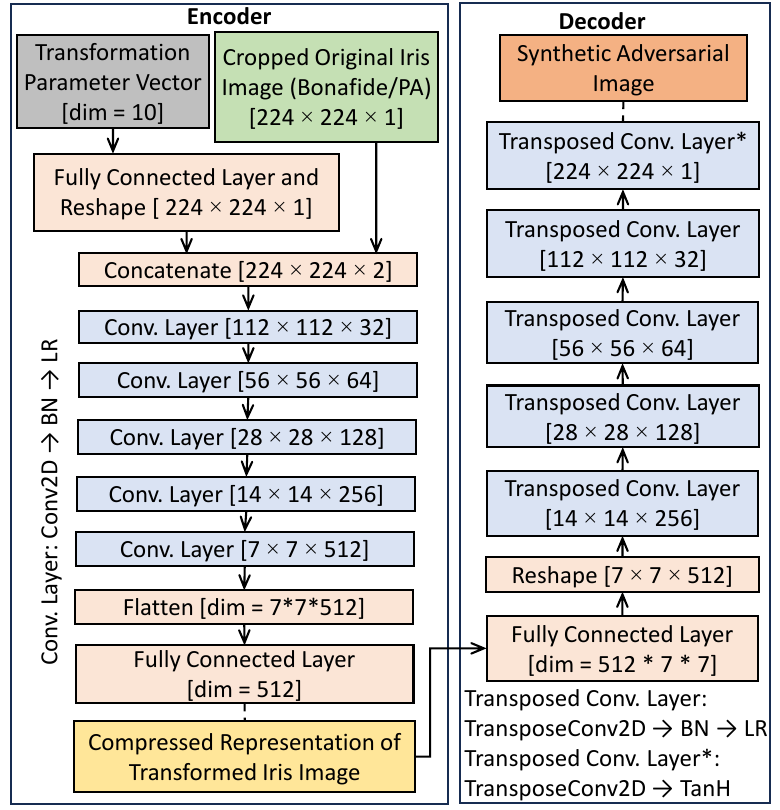}
		}}
  \end{center}
\vspace{-\baselineskip}
		\caption{Convolution Autoencoder (CAE) of ADV-GEN (BN: Batch normalization, LR: LeakyReLU activation). This inputs an image and a transformation parameter vector to generate a transformed image, unlike traditional autoencoders that input only the image and produce a reconstructed image.}
		\label{fig:autoencoder}
  \end{figure}

\begin{figure}[t]
  \begin{center}
      
		\fbox{
			\parbox[][0.23\textheight][c]{0.40\textwidth}{
				\includegraphics[width=0.40\textwidth, height=0.23\textheight]{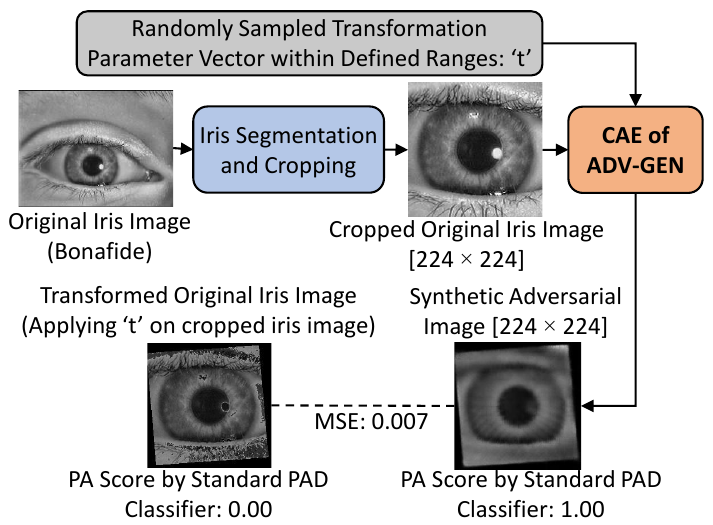}
		}}
    \end{center}
\vspace{-\baselineskip}
		\caption{Adversarial image generation with the trained CAE of ADV-GEN during inference.}
		\label{fig:inference}
  \vspace{-\baselineskip}
  \end{figure}

The ADV-GEN employs a CAE, with the encoder utilizing strided convolutions for downsampling and the decoder using transposed convolutions for upsampling. Figure \ref{fig:autoencoder} describes our encoder and decoder architectures. A normalized transformation parameter vector (scaled between 0 and 1 using defined ranges), is passed through a fully connected layer and then added as an additional channel to the original image to feed into the encoder. Input image pixels are normalized to the range $[-1,+1]$. We utilize Adam optimizer with a learning rate of 0.0001 to train the network for 500 epochs. Figure \ref{fig:inference} depicts the adversarial image generation process with the trained CAE.
\vspace{-\baselineskip}
\paragraph{\bf Why Use Transformation Parameters as Input?} Adversarial images are typically created by injecting carefully crafted perturbation noise into inputs, often derived using gradient-based techniques. Our approach aims to induce perturbations {\em conditioned} on transformation parameters to generate adversarial images. This method offers several advantages: (a) The transformed images increase intra-class variations among training samples. By using such images as ``seeds", a broad range of adversarial samples can be generated. (b) The nature of transformations considered in this work creates a smooth mapping between the input transformation parameter space and the output transformed image space (in contrast with using random noise as input along with a training sample). Further, the bounded transformation parameter space not only helps in generating semantically plausible adversarial images, but also accelerates the discovery process by constraining the search space. Thus, the transformation parameter vector helps in regularizing the adversarial generation process. This is borne out in our ablation study, where the exclusion of the input transformation parameter vector significantly degrades performance of the PAD classifier trained on such adversarial images.


\subsection{AA-PAD Classifier}
After generating adversarial images corresponding to each original training image, we select a subset of these samples to augment the training dataset. Initially, we filter the adversarial images by choosing those with a {\it squared L2-norm} below a certain threshold. This threshold is determined from a histogram plot of the squared L2-norm values (MSE) between the synthetic adversarial images and the transformed original images, preserving sufficient samples from both classes to be used in the subsequent step. 

Next, we apply {\it K-means clustering} on the embedding space of the selected adversarial images from each class {\it separately}. The embeddings are derived from a pre-trained DenseNet-121 model. We utilize $50,176$-dimensional feature vector from a flattened $1024\times7\times7$ feature set extracted at the end of DenseNet-121 architecture. From each cluster, we select a specified number of samples closest to the centroid to retain an approximately equal number of adversarial bonafides and adversarial PAs for inclusion in the training dataset. By combining these two processes, we ensure that the selected adversarial images closely resemble the transformed original images while maintaining diversity in their embedding space. 

Finally, we re-train a PAD classifier similar to the Standard PAD, but with the original training images augmented with the selected set of synthetic adversarial images, creating the AA-PAD classifier.  

\section{Experiments and Evaluation}
\label{sec:Experiments}
\subsection{Datasets}
We utilize the LivDet-Iris 2017 database \cite{Yambay2017}, comprising four datasets, in our experiments.
\vspace{-1.0\baselineskip}
\paragraph{\textbf{Clarkson Dataset (Cross-PA):}} All images were captured using the LG IrisAccess EOU2200 sensor. The training set includes bonafides, printouts of bonafide NIR images and 15 patterned contact types. The test set comprises additional PA types including 5 patterned contact lenses and visible light image printouts captured with an iPhone 5. 
\vspace{-1.0\baselineskip}
\paragraph{\textbf{Warsaw Dataset (Cross-sensor):}} It includes bonafides and their corresponding paper printouts, split into subject-disjoint training and test sets (subjects present in training set do not appear in the test set). The test set consists of two subject-disjoint subsets: known and unknown spoofs (PAs). All images in training set and known test set were captured by IrisGuard AD 100 sensor, while images in unknown test set by different sensors such as Aritech ARX-3M3C camera with SONY EX-View CCD sensor, Fujinon DV10X7.5A-SA2 lens and B+W 092 NIR filter.     
\vspace{-1.0\baselineskip}
\paragraph{\textbf{NotreDame Dataset (Cross-PA):}} It comprises bonafides and textured (cosmetic) contact lenses, divided into training and test sets. The test set was further split into known spoofs and unknown spoofs. Textured contact lenses in the training and known spoof sets were manufactured by Ciba, UCL and ClearLab, while the unknown spoof set includes lenses from Cooper and J\&J.
\vspace{-1.0\baselineskip}
\paragraph{\textbf{IIITD-WVU Dataset (Cross-database):}} The training set corresponds to IIITD database, including bonafides and textured contact lenses acquired in controlled settings using Cogent CIS 202 and VistaFA2E sensors, along with print attack images. The test set corresponds to the WVU database, comprising bonafides and textured contact lenses captured by the IriShield MK2120U mobile iris sensor under varying environmental conditions, as well as print attack images produced using two different printers.
 
\begin{figure*}
\begin{center}
    		\fbox{
			\parbox[][0.08\textheight][c]{0.96\linewidth}{
				\includegraphics[width=0.96\textwidth, height=0.08\textheight]{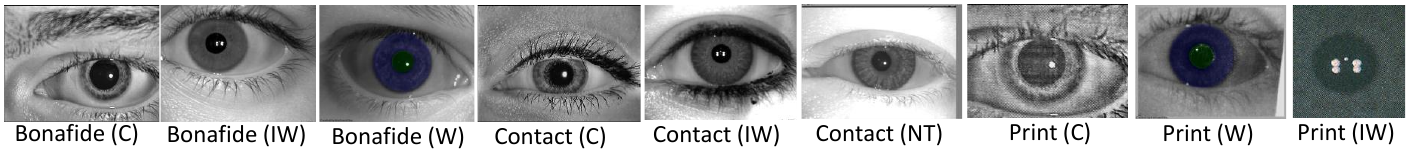}
		}}

\end{center}
\vspace{-\baselineskip}
\caption{Sample iris images from LivDet-Iris 2017 datasets (C:Clarkson, W:Warsaw, NT: NotreDame, IW: IIITD-WVU)}
		\label{fig:iris_samples}
  \vspace{-\baselineskip}
  \end{figure*}

Figure \ref{fig:iris_samples} shows sample iris images from this database. Table \ref{table:sample_numbers} lists the number of images that were eventually used, as the VeriEye iris-segmenter fails to segment certain irides. 

\begin{table}[t]
    \centering     
    \caption{Number of images retained after applying VeriEye iris-segmenter on the LivDet-Iris 2017 database. (Different from the original paper \cite{Yambay2017} due to segmentation failure for certain images.)}
    \vspace{-0.5\baselineskip}
    \begin{tabular}{|>{\centering\arraybackslash}m{1.5cm}|>{\centering\arraybackslash}m{0.8cm}|>{\centering\arraybackslash}m{0.8cm}|>{\centering\arraybackslash}m{0.8cm}|>{\centering\arraybackslash}m{0.8cm}|>{\centering\arraybackslash}m{0.8cm}|}
         \hline
         {\cellcolor{lightgray}\bf Dataset} & \multicolumn{2}{c|}{\cellcolor{lightgray}\textbf{Clarkson}} & \multicolumn{3}{c|}{\cellcolor{lightgray}\textbf{Warsaw}} \\
         \thickhline
         {\bf Split} & {\bf Train} & {\bf Test} & {\bf Train} & {\bf K.Test}& {\bf U.Test} \\
         \hline
         {\bf \#Bonafides} & 2,457 & 1,478 & 1,843 & 971 & 2,268  \\
         \hline
         {\bf \#PAs} & 1,944 & 1,434 & 2,664 & 2,011 & 2,107  \\
         \hline
         {\cellcolor{lightgray}\bf Dataset} & \multicolumn{3}{c|}{\cellcolor{lightgray}\textbf{NotreDame}} & \multicolumn{2}{c|}{\cellcolor{lightgray}\textbf{IIITD-WVU}} \\
         \thickhline
         {\bf Split} & {\bf Train} & {\bf K.Test} & {\bf U.Test} & {\bf Train} & {\bf Test}\\
         \hline
          {\bf \#Bonafides} & 599 & 899 & 896 & 2,155 & 700\\
          \hline
         {\bf \#PAs} & 599 & 889 & 900 & 2,045 & 3,424\\
         \hline
    \end{tabular}
    \label{table:sample_numbers}
    \begin{tablenotes}
       \item [a] \footnotesize{K.- Known; U.- Unknown}
     \end{tablenotes}
    \vspace{-\baselineskip}
\end{table}

\subsection{Evaluation Protocol}

To compare the performance of AA-PAD against the Standard PAD classifier, we compute True Detection Rate (TDR) at a False Detection Rate (FDR) of 0.1\%. TDR indicates the percentage of PA samples correctly classified by the classifier, while FDR denotes the percentage of bonafide samples incorrectly detected as PAs. 

While comparing performance with existing literature, we utilize APCER (Attack Presentation Classification Error Rate), BPCER (Bonafide Presentation Classification Error Rate) and HTER (Half Total Error Rate) metrics. APCER is the percentage of PA samples classified as bonafide, equivalent to (1-TDR), and BPCER is the same as FDR. HTER is the average of APCER and BPCER.  

\subsection{Experimental Results}

\begin{figure}[t]
  \begin{center}
		\fbox{
			\parbox[][0.16\textheight][c]{0.47\textwidth}{\vspace{-2.2\baselineskip}
				\includegraphics[width=0.47\textwidth, height=0.17\textheight]{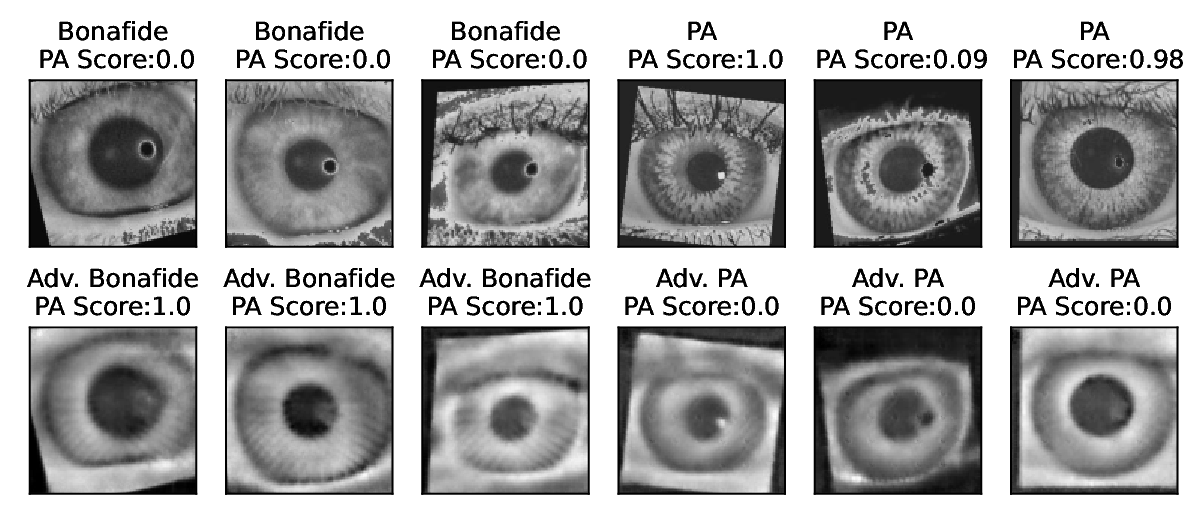}\\\\[-1.8 cm]
		}}
    \end{center}
\vspace{-\baselineskip}
		\caption{Samples of transformed original iris images (top) and synthetic adversarial images generated using ADV-GEN (bottom) with the Clarkson training set. PA score is obtained by the Standard PAD classifier.}
		\label{fig:generated_samples}
  \vspace{-\baselineskip}
  \end{figure}

We implement our proposed method using the Clarkson, IIITD-WVU, and NotreDame training sets {\it separately}, considering each dataset as an individual domain. Figure \ref{fig:generated_samples}  displays examples of synthetic adversarial images generated by the ADV-GEN. The synthetic adversarial images are filtered by selecting those with a {\bf squared L2 norm less than a threshold of 0.01} for all experiments, followed by K-means clustering applied to the embedding space of the selected images. The number of clusters ($K$) and the selected images ($S$) per cluster were determined through 
a trial-and-error approach, informed by the size of training dataset.
\vspace{-1.0\baselineskip}
\paragraph{\textbf{Experiment A (Clarkson):}} We set $K$ to 10 and $S$ to 20, selecting 200 adversarial images from each class, which equals 9.1\% of the original training images.
\vspace{-1.0\baselineskip}
\paragraph{\textbf{Experiment B (IIITD-WVU):}} Using the same parameters as in Exp. A, we select 200 adversarial images for each class, constituting 9.5\% of the original training images.
\vspace{-1.0\baselineskip}
\paragraph{\textbf{Experiment C (NotreDame):}} Due to the smaller training set in this case, we set the parameters to $K=5$ and $S=25$, selecting 125 adversarial images from each class, which equals ~21\% of the original training images. 

Next, the AA-PAD classifier is trained on each dataset using the original images, augmented with selected synthetic adversarial images, and evaluated across all four datasets to assess both intra- and cross-dataset performance.
  
\begin{figure}[t]
  \begin{center}
		\fbox{
			\parbox[][0.16\textheight][c]{0.47\textwidth}{\vspace{-2.6\baselineskip}
				\includegraphics[width=0.47\textwidth, height=0.18\textheight]{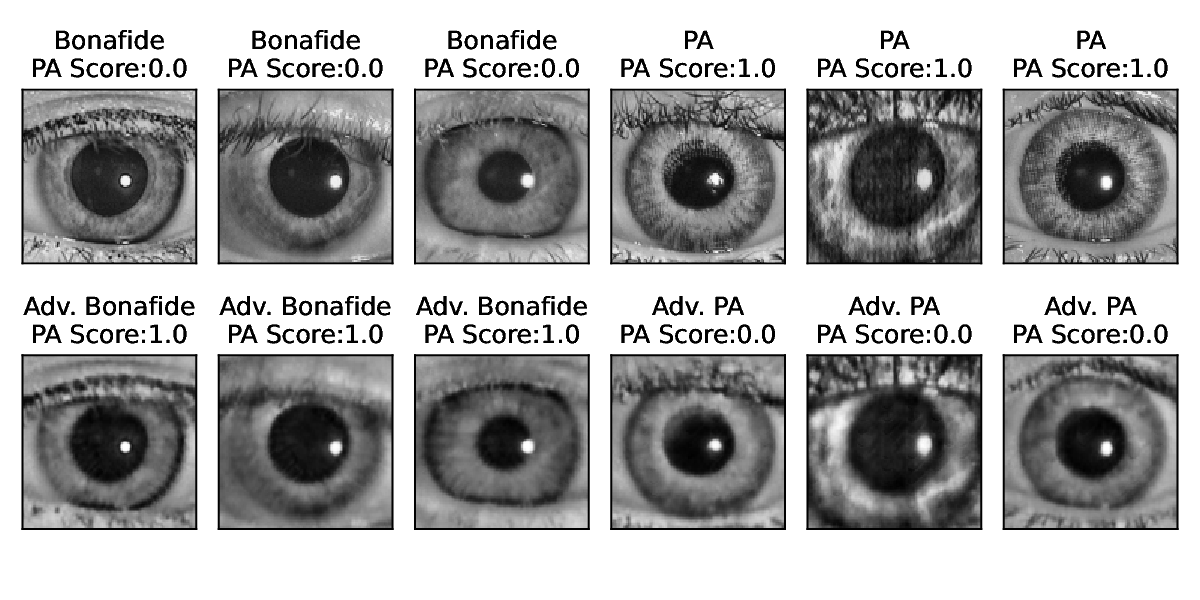}\\[-1.8 cm]
		}}
    \end{center}
\vspace{-\baselineskip}
		\caption{Samples of original images (top) and synthetic adversarial images using ADV-GEN without transformation parameters (bottom) with the Clarkson training set. PA score is obtained by the Standard PAD classifier.}
		\label{fig:generated_samples_wo_transform}
  \vspace{-0.5\baselineskip}
  \end{figure}

\vspace{-\baselineskip}  
\paragraph{\bf Ablation study on AA-PAD without using ADV-GEN:} We also conducted an ablation study on AA-PAD classifier by augmenting the adversarial images, produced by applying the same transformation parameters that were used during inference with the ADV-GEN, to the original images. We observe that only a few {\bf transformed original iris images} were classified as adversarial by the Standard PAD classifier: 
\begin{itemize}[topsep=0.1em, noitemsep,leftmargin=*]
    \item {\bf Exp. A:} Adversarial bonafides:$\sim$11.5\% and PAs:$\sim$1.5\%.
    \item {\bf Exp. B:} Adversarial bonafides:$\sim$1.5\% and PAs:$\sim$1\%.
    \item {\bf Exp. C:} Adversarial bonafides:$\sim$0\% and PAs:$\sim$17\%.
\end{itemize} 
\vspace{-\baselineskip}
\paragraph{\bf Ablation Study on AA-PAD without Transformation Parameters in ADV-GEN:} To assess the impact of using transformation parameters, we train ADV-GEN without incorporating them, and using only the original images. Figure \ref{fig:generated_samples_wo_transform} displays samples of generated adversarial images. Although the generated images are adversarial with respect to the Standard PAD classifier, their inclusion during AA-PAD training with identical selection parameters, did not improve cross-dataset performance. This suggests that the transformation parameters do play a crucial role in enabling the network to learn domain-invariant features.   

\vspace{-\baselineskip}
\paragraph{\bf Discussion:} Table \ref{table:results} reports both intra- and cross-dataset performance of Standard PAD and AA-PAD classifiers, along with both the ablation study results on the AA-PAD classifier. We observe that in Exp. A and B, the AA-PAD effectively improves TDR\%@0.1\% FDR compared to Standard PAD classifier. However, in Exp. C, the AA-PAD could not beat the performance of Standard PAD in most cases, potentially due to the smaller number of training images in the NotreDame dataset, which may have affected the quality of the generated adversarial images. Further, the ablation studies on the AA-PAD classifier highlight the effectiveness of generating adversarial images using ADV-GEN as well as the benefits of using the transformation parameters. The Warsaw dataset, containing only print attack images, is excluded from training as it did not significantly contribute for PAD generalization towards cosmetic contact lenses. Literature also suggests that print images are easily detected, highlighting the challenge of generalizing PAD methods trained on such datasets to other types of attacks. 
\begin{table}
    \centering
    \caption{Intra- and cross-dataset evaluation (TDR\%@0.1\%FDR) of {\bf (1)} Standard PAD classifier; {\bf (2)} Ablation study on AA-PAD without ADV-GEN; {\bf (3)} Ablation study on AA-PAD without using transformation parameters in ADV-GEN; {\bf (4)} AA-PAD classifier} 
    \vspace{-0.5\baselineskip}
    \begin{tabular}{|>{\centering\arraybackslash}p{0.24cm}|>{\centering\arraybackslash}p{1.3cm}|>{\centering\arraybackslash}p{0.8cm}|>{\centering\arraybackslash}p{0.8cm}|>{\centering\arraybackslash}p{0.8cm}|>{\centering\arraybackslash}p{0.8cm}|>{\centering\arraybackslash}p{0.6cm}|}
    \hline
         
          \cellcolor{lightgray}& \cellcolor{lightgray}\textbf{Clarkson} & \multicolumn{2}{c|}{\cellcolor{lightgray}\textbf{Warsaw}} & \multicolumn{2}{c|}{\cellcolor{lightgray}\textbf{NotreDame}} & \cellcolor{lightgray}\textbf{IW} \\
         \hline
         \cellcolor{lightgray}& \cellcolor{lightgray}\textbf{Test} & \cellcolor{lightgray}{\bf K.Test}& \cellcolor{lightgray}{\bf U.Test} & \cellcolor{lightgray}{\bf K.Test} & \cellcolor{lightgray}{\bf U.Test} & \cellcolor{lightgray}{\bf Test}\\
            \hline
            \cellcolor{lightgray}& \multicolumn{6}{c|}{\textbf{\cellcolor{lightblue}Exp. A: Training Dataset - Clarkson}}\\
         \thickhline
          \cellcolor{lightgray}{\bf (1)} & 89.1 & 36.3 & 40.7 & 63.8 & 21.4 & 44.2\\
            \hline
           
           \cellcolor{lightgray}{\bf (2)} & 88.3 & 61.1 & 35.0 & 85.4 & 36.7 & 48.4\\
           \hline
           \cellcolor{lightgray}{\bf (3)} & 89.6 & 25.1 & 32.7 & 75.1 & 27.3 & 33.5\\
         \hline
         \cellcolor{lightgray}{\bf (4)} & {\bf 91.1} & {\bf 92.4} & {\bf 88.7} & {\bf 94.3} & {\bf 49.9} & {\bf 51.8}\\
           \hline
         \cellcolor{lightgray}&\multicolumn{6}{c|}{\cellcolor{lightblue}\textbf{Exp. B: Training Dataset - IIITD-WVU (IW)}}\\
        \thickhline
         \cellcolor{lightgray}{\bf (1)} & 23.7 & 58.2 & {\bf 97.3} & 93.1 & 49.9 & 45.9\\
            \hline
          \cellcolor{lightgray}{\bf (2)} & 22.5 & 52.2 & 68.7 & 90.9 & 36.9 & 54.4\\
         \hline
          \cellcolor{lightgray}{\bf (3)}& 18.1 & 81.6 & 73.1 & 98.2 & 37.7 & 64.9\\
            \hline
            \cellcolor{lightgray}{\bf (4)} & {\bf51.5} & {\bf 85.3} & 89.9 & {\bf 98.5} & {\bf 57.3} & {\bf 68.0}\\
           \hline
         \cellcolor{lightgray}&\multicolumn{6}{c|}{\cellcolor{lightblue}\textbf{Exp. C: Training Dataset - NotreDame}}\\
        \thickhline
         \cellcolor{lightgray}{\bf (1)} & 56.8 & {\bf 100.0} & {\bf 94.9} & {\bf 100.0} & {\bf 93.8} & {\bf 87.0}\\
            \hline
          \cellcolor{lightgray}{\bf (2)} & 70.1 & 98.2 & 91.8 & 99.9 & 92.7 & 74.9\\
         \hline
          \cellcolor{lightgray}{\bf (3)}& 47.3 & 82.6 & 35.9 & {\bf 100.0} & 43.7 & 54.9\\
            \hline
            \cellcolor{lightgray}{\bf (4)} & {\bf76.5} & {\bf 100.0} & 87.9 & {\bf 100.0} & 75.6 & 76.0\\
           \hline
          
    \end{tabular}
    \vspace{-\baselineskip}
    \label{table:results}
\end{table}
\vspace{-\baselineskip}
\paragraph{\bf Evaluation with other architectures:} We justify our choice of using the DenseNet architecture by testing our approach using other architectures (Table \ref{table:results_other}), focusing on the Clarkson training set due to its larger sample size and strong intra-dataset performance. While the AA-PAD classifier demonstrated improved performance in some cases where the Standard PAD struggled, the gains were not as consistent as those achieved with the DenseNet architecture. Previous studies \cite{Yadav2019,Sharma2020} have similarly emphasized the benefits of using DenseNet in achieving high accuracy for iris PAD.

\begin{table}[!h]
    \centering
    \caption{Intra- and cross-dataset evaluation (TDR\%@0.1\%FDR) of our method with other architectures, viz., VGG-19 and ResNet-50, using the Clarkson training set: {\bf (a)} Standard PAD classifier; {\bf (b)} AA-PAD classifier}
    \vspace{-0.5\baselineskip}
    \begin{tabular}{|>{\centering\arraybackslash}m{0.26cm}|>{\centering\arraybackslash}p{1.3cm}|>{\centering\arraybackslash}p{0.82cm}|>{\centering\arraybackslash}p{0.82cm}|>{\centering\arraybackslash}p{0.82cm}|>{\centering\arraybackslash}p{0.82cm}|>{\centering\arraybackslash}p{0.5cm}|}
    \hline
          \cellcolor{lightgray}& \cellcolor{lightgray}\textbf{Clarkson} & \multicolumn{2}{c|}{\cellcolor{lightgray}\textbf{Warsaw}} & \multicolumn{2}{c|}{\cellcolor{lightgray}\textbf{NotreDame}} & \multicolumn{1}{c|}{\cellcolor{lightgray}\textbf{IW}} \\
         \hline
         \cellcolor{lightgray}& \cellcolor{lightgray}\textbf{Test} & \cellcolor{lightgray}{\bf K.Test}& \cellcolor{lightgray}{\bf U.Test} & \cellcolor{lightgray}{\bf K.Test} & \cellcolor{lightgray}{\bf U.Test} & \cellcolor{lightgray}{\bf Test}\\
            \hline
            \cellcolor{lightgray}& \multicolumn{6}{c|}{\textbf{\cellcolor{lightblue}VGG-19 \cite{Simonyan2015}}}\\
         \thickhline
          \cellcolor{lightgray}{\bf (a)} & \bf 85.5 & 55.2 & \bf 97.3 & \bf 98.7 & 55.9 & 62.3\\
            \hline
           
           \cellcolor{lightgray}{\bf (b)} & 83.4 &  \bf 95.8 & 88.4 & 97.9 & \bf 58.0 & \bf 68.6\\
           \hline 
           \cellcolor{lightgray}&\multicolumn{6}{c|}{\cellcolor{lightblue}\textbf{ResNet-50 \cite{He2016}}}\\
        \thickhline
         \cellcolor{lightgray}{\bf (a)} & \bf 91.8 & 82.1 & \bf 86.9 & \bf 82.3 & 30.2 & \bf 58.0\\
            \hline
          \cellcolor{lightgray}{\bf (b)} & 90.5 & \bf 88.9 & 78.2 & 79.9 & \bf 41.0 & 52.5\\
         \hline
    \end{tabular}
    \vspace{-\baselineskip}
    \label{table:results_other}
\end{table}

\subsection{Comparison with Existing PAD Algorithms} We compare the performance of our AA-PAD classifier with existing algorithms \cite{Fang2022, Fang2023, Li2023}, on which the same cross-dataset experiments were conducted using LivDet-Iris 2017 (Table \ref{table:comparison}). The AA-PAD classifier achieves lower APCER and HTER in all cases except one. Notably, while Exp. C (NotreDame training) does not outperform the Standard PAD classifier in terms of TDR\%@0.1\% FDR (Table \ref{table:results}), it still surpasses the existing methods in cross-dataset scenarios when evaluated using error rate metrics. Since this dataset includes only textured contact lenses, this experiment partially accounts for unknown PAs (print images) when tested on the Clarkson and IIITD-WVU datasets.

\begin{table}[t]
\begin{threeparttable}[t]
    \centering
    \vspace{-0.5\baselineskip}
    \caption{Cross-dataset performance comparison of the AA-PAD classifier on the LivDet-Iris 2017 datasets}
    \vspace{-0.5\baselineskip}
    \begin{tabular}{|>{\centering\arraybackslash}m{1.3cm}|p{1.65cm}|>{\centering\arraybackslash}p{1.1cm}|>{\centering\arraybackslash}p{1.1cm}|>{\centering\arraybackslash}p{0.9cm}|}
        \hline
         \cellcolor{lightgray}\textbf{Test}  &\cellcolor{lightgray}{\bf Classifier} & \cellcolor{lightgray}{\bf APCER} & \cellcolor{lightgray}{\bf BPCER} & \cellcolor{lightgray}{\bf HTER}\\
        \cline{2-5}
         \cellcolor{lightgray}\textbf{Dataset} & \multicolumn{4}{c|}{\cellcolor{lightblue}\textbf{Exp. A: Training Dataset - Clarkson}} \\
         \thickhline
         \multirow{3}{1.3cm}{\textbf{Notre-Dame}} &DA \cite{Fang2022}\tnote{a} & - & - & {\bf 7.89} \\
         \cline{2-5}
         &PBS \cite{Fang2023} & 64.83 & {\bf 0.00} & 32.42 \\
         \cline{2-5}
         &A-PBS \cite{Fang2023} & 46.16 & {\bf 0.00} & 23.08 \\
         \cline{2-5}
         &SDGG \cite{Li2023} & - & - & 8.47 \\
         \cline{2-5}
         &\textcolor{blue}{\bf AA-PAD} & {\bf 23.18} & 0.56 & 11.86 \\
        \hline
           \multirow{3}{1.3cm}{\textbf{IIITD-WVU}} &DA \cite{Fang2022}\tnote{a} & - & - & 19.71\\
           \cline{2-5}
           &PBS \cite{Fang2023} & 84.97 & {\bf 0.00} & 42.48\\
           \cline{2-5}
           &A-PBS \cite{Fang2023} & 68.34 & {\bf 0.00} & 34.17\\
           \cline{2-5}
           &SDGG \cite{Li2023} & - & - & 16.68 \\
         \cline{2-5}
           &\textcolor{blue}{\bf AA-PAD} & {\bf 26.84} & 4.57 & {\bf 15.71}\\
           \thickhline
          \multirow{4}{1.3cm}{\textbf{Clarkson}} & \multicolumn{4}{c|}{\cellcolor{lightblue} \textbf{Exp. B: Training Dataset - IIITD-WVU}}\\
          \cline{2-5}
          &DA \cite{Fang2022}\tnote{a} & - & - & 18.29 \\
         \cline{2-5}
          &PBS \cite{Fang2023} & 47.17 & {\bf 0.00} & 94.34 \\
         \cline{2-5}
         &A-PBS \cite{Fang2023} & 20.80 & 32.10 & 21.95 \\
         \cline{2-5}
         &SDGG \cite{Li2023} & - & - & 19.36 \\
         \cline{2-5}
         &\textcolor{blue}{\bf AA-PAD} & {\bf 14.16 } & 8.32 & {\bf 11.24} \\
        \hline
           \multirow{3}{1.3cm}{\textbf{Notre-Dame}} &DA \cite{Fang2022}\tnote{a} & - & - & 7.36\\
           \cline{2-5}
           &PBS \cite{Fang2023} & 33.33 & 0.39 & 16.86\\
           \cline{2-5}
           &A-PBS \cite{Fang2023} & 54.72 & {\bf 0.33} & 27.61\\
           \cline{2-5}
           &SDGG \cite{Li2023} & - & - & 6.03 \\
         \cline{2-5}
           &\textcolor{blue}{\bf AA-PAD} & {\bf 5.84} & 1.90 & {\bf 3.87}\\
           \thickhline
           \multirow{4}{1.3cm}{\textbf{Clarkson}} & \multicolumn{4}{c|}{\cellcolor{lightblue} 
 \textbf{Exp. C: Training Dataset - NotreDame}}\\
          \cline{2-5}
          &DA \cite{Fang2022}\tnote{a} & - & - & 10.58 \\
          \cline{2-5}
          &PBS \cite{Fang2023} & 28.15 & 66.40 & 47.28 \\
         \cline{2-5}
         &A-PBS \cite{Fang2023} & 14.76 & 31.72 & 23.24 \\
         \cline{2-5}
         &SDGG \cite{Li2023} & - & - & 10.20 \\
         \cline{2-5}
         &\textcolor{blue}{\bf AA-PAD} & {\bf 8.86 } & {\bf 4.06} & {\bf 6.46} \\
         \hline
           \multirow{3}{1.3cm}{\textbf{IIITD-WVU}} &DA \cite{Fang2022}\tnote{a} & - & - & 8.82\\
           \cline{2-5}
           &PBS \cite{Fang2023} & 22.24 & 6.83 & 14.54\\
           \cline{2-5}
           &A-PBS \cite{Fang2023} & 15.34 & {\bf 2.56} & 8.95\\
           \cline{2-5}
           &SDGG \cite{Li2023} & - & - & 16.69 \\
         \cline{2-5}
           &\textcolor{blue}{\bf AA-PAD} & {\bf 10.89} & 2.71 & {\bf 6.80}\\
           \hline
         
    \end{tabular}
    \label{table:comparison}
    \begin{tablenotes}
       \item [a] \footnotesize{Considered best performing model}
     \end{tablenotes}
\end{threeparttable}
\end{table}

Further, the IIITD-WVU dataset poses a challenging cross-dataset scenario where existing PAD algorithms have shown subpar performance. Our AA-PAD classifier shows better performance with APCER of 2.28\% and HTER of 3.43\% (Table \ref{table:comaparison_IW}). It also outperforms the top-performing models from the LivDet-Iris 2020 competition \cite{Das2020}, achieving 15.98\% APCER and 11.22\% HTER (Table \ref{table:comaparison_livdet2020}).

\begin{table}[t]
    \centering
    \vspace{-0.5\baselineskip}
    \caption{Comparison of AA-PAD classifier with other methods on the IIITD-WVU dataset representing a cross-database evaluation scenario (Training: IIITD database, Test: WVU database). ``Winner'' refers to LivDet-Iris 2017 winner for this dataset.}
    \vspace{-0.5\baselineskip}
    \begin{tabular}{|p{2.6cm}|>{\centering\arraybackslash}m{1.2cm}|>{\centering\arraybackslash}m{1.2cm}|>{\centering\arraybackslash}m{1.2cm}|}
         \hline
         {\cellcolor{lightgray}\bf Classifier} & {\cellcolor{lightgray}\bf APCER} & {\cellcolor{lightgray}\bf BPCER} & {\cellcolor{lightgray}\bf HTER}\\
         \thickhline
         Winner \cite{Yambay2017} & 29.40 & {\bf 3.99} & 16.70\\ \hline
         Meta-Fusion \cite{Kuehlkamp2019} & 12.32 & 17.52 & 14.92\\ \hline
         D-NetPAD \cite{Sharma2020} & 36.41 & 10.12 & 23.27\\ \hline
         MSA \cite{Fang2021MSA} & 2.31  & 19.94 & 11.13\\ \hline
         PBS \cite{Fang2023} & 5.76 & 8.26 & 7.01\\ \hline
         A-PBS \cite{Fang2023}& 8.86 & 4.13 & 6.50\\ \hline
         \textcolor{blue}{\bf AA-PAD (Ours)} &{\bf 2.28} & 4.57 & {\bf 3.43}\\
        \hline
    \end{tabular}
    \label{table:comaparison_IW}
    \vspace{-\baselineskip}
\end{table}


\begin{table}[t]
    \centering
    \vspace{-\baselineskip}
    \caption{Comparison of AA-PAD classifiers trained on different LivDet-Iris 2017 datasets (C: Clarkson; IW: IIITD-WVU; NT: NotreDame) and tested on the LivDet-Iris 2020 test dataset}
    \begin{tabular}{|p{1cm}|>{\centering\arraybackslash}p{0.7cm}|>{\centering\arraybackslash}p{0.7cm}|>{\centering\arraybackslash}p{0.7cm}|>{\centering\arraybackslash}p{0.7cm}|>{\centering\arraybackslash}p{0.7cm}|>{\centering\arraybackslash}p{0.7cm}|}
         \hline
         \cellcolor{lightgray} \textbf{Metrics} & \multicolumn{3}{c|}{\cellcolor{lightgray}\bf LivDet-Iris 2020 } & \multicolumn{3}{c|}{\cellcolor{lightgray}\bf AA-PAD Classifiers}\\
         \cline{5-7}
         \cellcolor{lightgray}& \multicolumn{3}{c|}{\cellcolor{lightgray}\bf Submissions \cite{Das2020}} & {\cellcolor{lightgray}\bf C} & \cellcolor{lightgray}\bf IW & \cellcolor{lightgray}\bf NT \\
         \thickhline
         APCER & 59.1 &48.68 &57.8 & 48.04  &27.77  &{\bf 15.98}\\
         \hline
         BPCER & {\bf 0.46} &11.59 &40.31 & 1.68 &5.31 &6.46\\
         \hline
         HTER & 29.78 &30.14& 49.06& 24.86& 16.54& {\bf 11.22}\\
        \hline
    \end{tabular}
    \label{table:comaparison_livdet2020}
\end{table}

\subsection{Visualization and Explainability}

\begin{figure}[t]
    \centering
    \begin{subfigure}[b]{0.23\textwidth}
      \begin{center}
		\fbox{
			\parbox[][0.16\textheight][c]{0.92\textwidth}{
        \includegraphics[width=0.92\textwidth, height = 0.16\textheight]{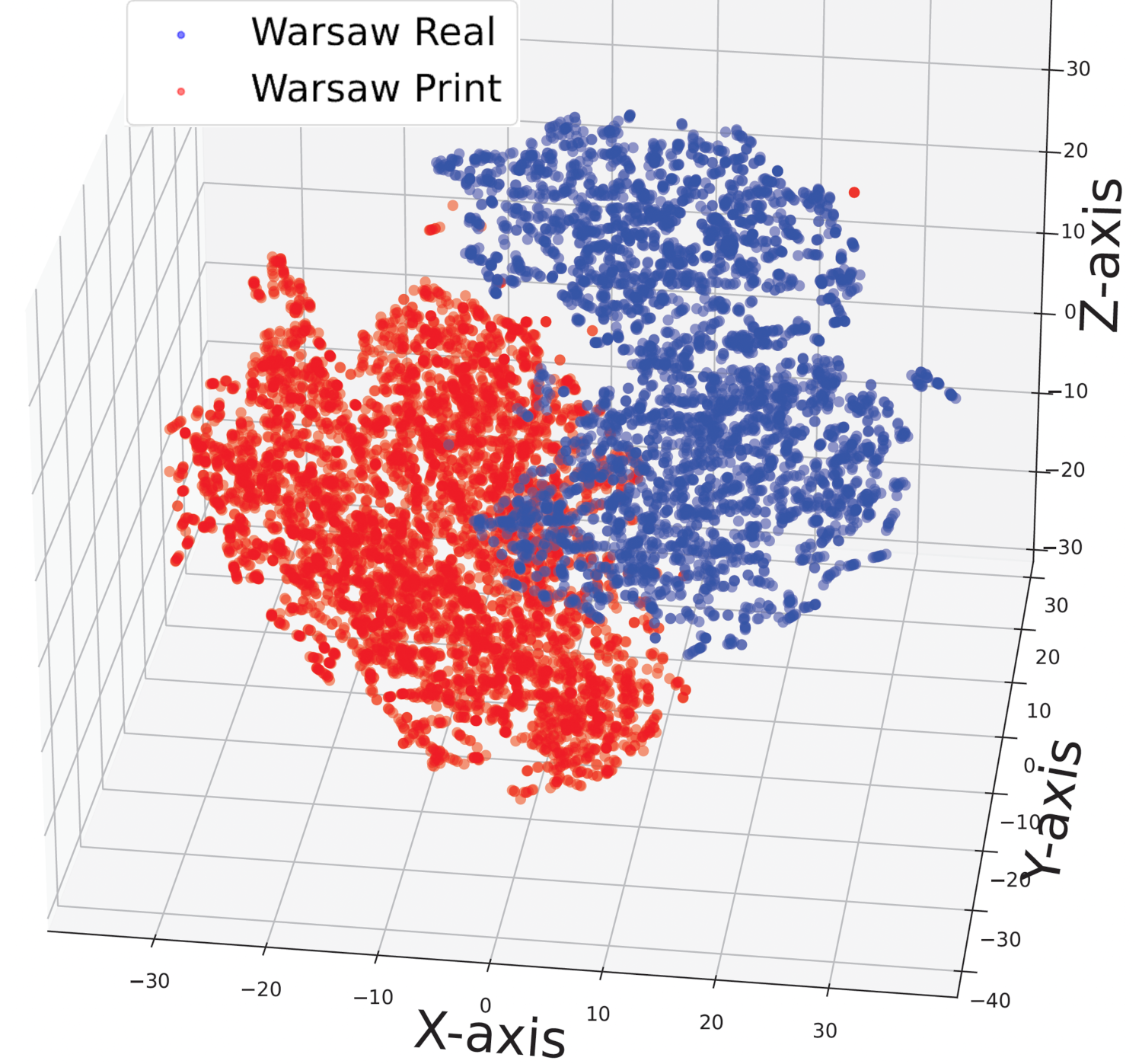}
        }}
        \end{center}
        \vspace{-\baselineskip}
        \caption{Standard PAD Classifier}
        \label{fig:kde-1}
    \end{subfigure}%
    \medskip
    \begin{subfigure}[b]{0.23\textwidth}
        \begin{center}
		\fbox{
			\parbox[][0.16\textheight][c]{0.92\textwidth}{
        \includegraphics[width=0.92\textwidth, height= 0.16\textheight]{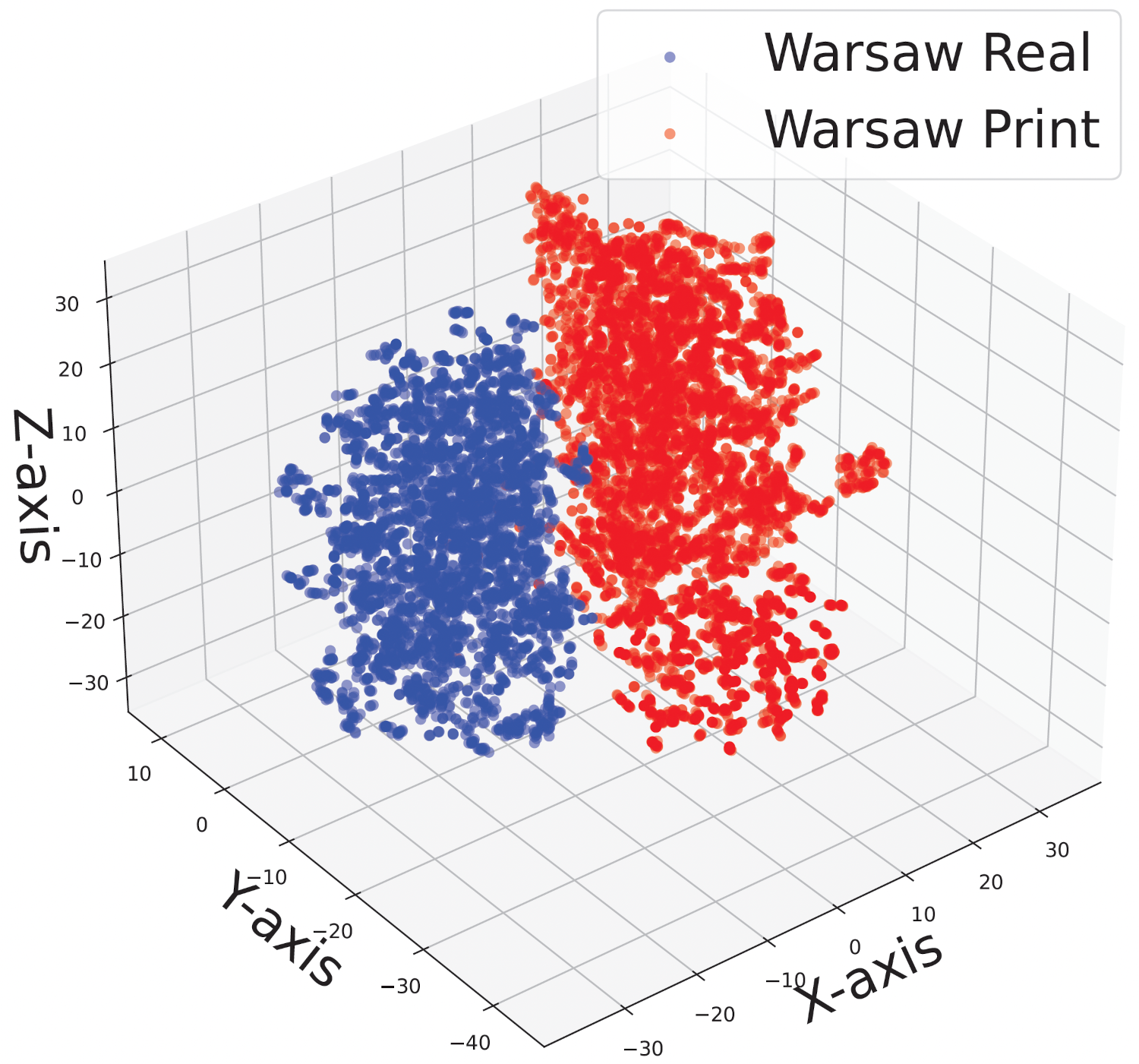}
        }}
        \end{center}
        \vspace{-\baselineskip}
        \caption{AA-PAD Classifier}
        \label{fig:kde-2}
    \end{subfigure}
    \centering
    \vspace{-\baselineskip}
    \caption{t-SNE feature distributions of all samples in the Warsaw test set using Clarkson-trained classifiers (Exp. A). Standard PAD projects the two classes in a feature space that needs a complex, higher-order boundary function, while AA-PAD projects them in a feature space that can be separated by an approximately linear boundary, leading to improved classification performance.}
    \label{fig:kde-warsaw}
 \vspace{-0.5\baselineskip}
\end{figure}

\begin{figure}[t]
    \begin{center}
		\fbox{
			\parbox[][0.07\textheight][c]{0.47\textwidth}{\vspace{-1.8\baselineskip} 
	\includegraphics[width=0.47\textwidth, height=0.12\textheight]{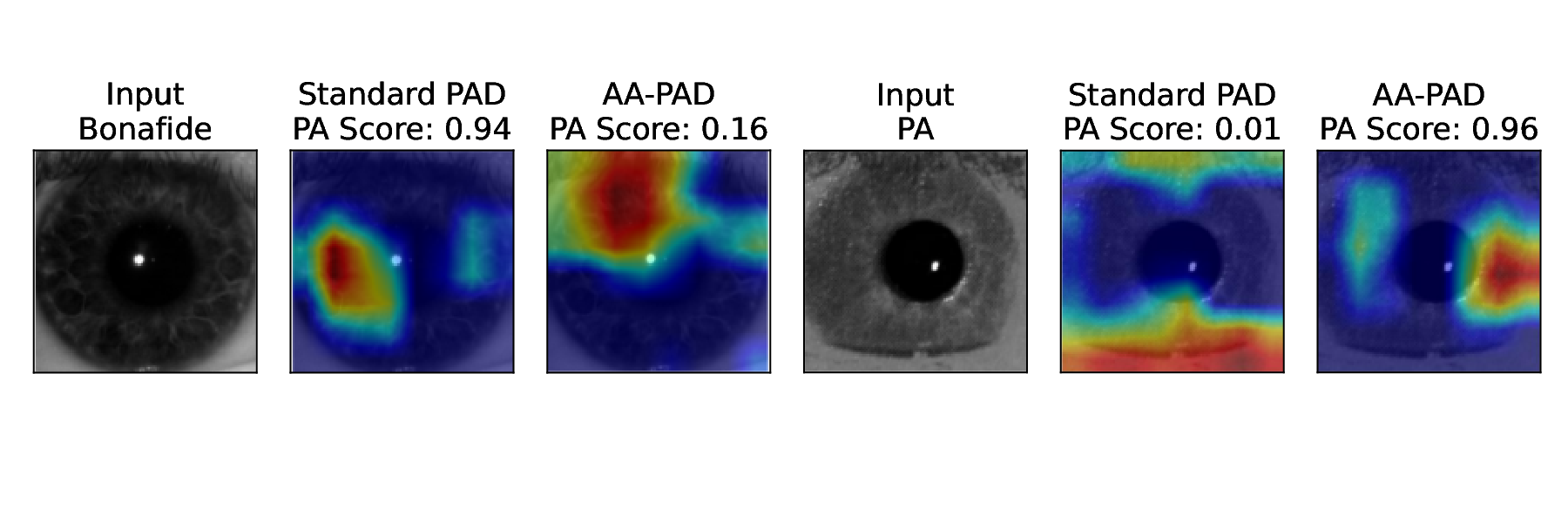}\\[-1.5 cm]
		}}
    \end{center}
    \vspace{-\baselineskip}
    \caption{Saliency maps (generated by GradCAM++ \cite{gradcam++}) on Warsaw test image samples by the Clarkson-trained classifiers (Exp. A), describing the models' focus during classification.}
    \label{fig:gradcam++}
  \vspace{-\baselineskip}
\end{figure}

We evaluate the extracted features for each class using the Clarkson-trained classifiers (Exp. A) on the Warsaw test set. Figure \ref{fig:kde-warsaw} shows t-SNE \cite{van2008visualizing-tsne} distribution plots of these features. The $1024\times7\times7$ feature maps obtained from DenseNet architecture are passed through an average pooling layer, producing $1024$-dimensional vector, which is then reduced to a $3$ dimensions for visualization. To further investigate model attention during decision-making, we examine the saliency maps of Warsaw unknown test cases, generated using GradCAM++~\cite{gradcam++} (Figure \ref{fig:gradcam++}) for Clarkson-trained classifiers. We choose GradCAM++ for its class invariance property (as our PAD classifier outputs a single score), and its independence from gradient backpropagation. Upon analyzing the intersection between the top $50\%$ region of the saliency map with the segmented iris area, we observe that the AA-PAD classifier's saliency map covers an average of $42.6\%$ of the iris area in the Warsaw unknown test set, compared to only $29.1\%$ by the Standard PAD.

\section{Conclusion}
\label{sec:Conclusion}
This work introduces an adversarial augmentation method to enhance the generalizability of an iris PAD classifier. We proposed a novel CAE model, ADV-GEN, which generates semantically meaningful adversarial images by leveraging transformation parameters from classical DA techniques. Experiments on the LivDet-Iris 2017 database and the LivDet-Iris 2020 dataset demonstrate the potential of this method in generalizing PA detection across datasets, sensors, PAs, and acquisition environments. Future directions include integrating advanced generative models and additional transformations, extending the method to other biometric modalities, and refining adversarial sample selection through advanced clustering techniques.



{\small
\balance
\bibliographystyle{ieee_fullname}
\bibliography{paper}
}

\end{document}